\documentclass[conference]{IEEEtran}
\IEEEoverridecommandlockouts
\usepackage{cite}
\usepackage{amsmath,amssymb,amsfonts}
\usepackage{algorithmic}
\usepackage{graphicx}
\usepackage{textcomp}
\usepackage{xcolor}
\usepackage{booktabs} 
\usepackage{tabulary}
\usepackage{amsmath,amsfonts,amssymb}
\usepackage{graphicx}
\usepackage{footnote}
\makesavenoteenv{tabular}
\makesavenoteenv{table}
\usepackage{subcaption}
\usepackage{enumitem}
\usepackage{romannum}
\usepackage{authblk}
\usepackage{hyperref}

\newtheorem{mydef}{Definition}
\newcommand*{\affaddr}[1]{#1} 
\newcommand*{\affmark}[1][*]{\textsuperscript{#1}}
\newcommand*{\email}[1]{\texttt{#1}}

\def\BibTeX{{\rm B\kern-.05em{\sc i\kern-.025em b}\kern-.08em
    T\kern-.1667em\lower.7ex\hbox{E}\kern-.125emX}}
\begin{document}

\title{JSCN: Joint Spectral Convolutional Network for \\ Cross Domain Recommendation}

\author{Zhiwei~Liu\affmark[1], Lei~Zheng\affmark[1], Jiawei~Zhang\affmark[2], Jiayu~Han\affmark[3], and Philip S.~Yu\affmark[1]\\
\affaddr{\affmark[1]Department of Computer Science, University of Illinois at Chicago, IL, USA} \\ \email{\{zliu213, lzheng21, psyu\}@uic.edu}\\
\affaddr{\affmark[2]IFM Lab, Department of Computer Science, Florida State University, FL, USA};
\email{jzhang@cs.fsu.edu} \\
\affaddr{\affmark[3]Department of Computer Science, Jilin University, Changchun, China}; 
\email{jyhan15@mails.jlu.edu.cn}
}


\maketitle

\begin{abstract}

Cross-domain recommendation can alleviate the data sparsity problem in recommender systems. To transfer the knowledge from one domain to another, one can either utilize the neighborhood information or learn a direct mapping function. However, all existing methods ignore the high-order connectivity information in cross-domain recommendation area and suffer from the domain-incompatibility problem. In this paper, we propose a \textbf{J}oint \textbf{S}pectral \textbf{C}onvolutional \textbf{N}etwork (JSCN) for cross-domain recommendation. JSCN will simultaneously operate multi-layer spectral convolutions on different graphs, and jointly learn a domain-invariant user representation with a domain adaptive user mapping module. As a result, the high-order comprehensive connectivity information can be extracted by the spectral convolutions and the information can be transferred across domains with the domain-invariant user mapping. The domain adaptive user mapping module can help the incompatible domains to transfer the knowledge across each other. Extensive experiments on $24$ Amazon rating datasets show the effectiveness of JSCN in the cross-domain recommendation, with $9.2\%$ improvement on recall and $36.4\%$ improvement on MAP compared with state-of-the-art methods. Our code is available online ~\footnote{https://github.com/JimLiu96/JSCN}.
\end{abstract}


\section{Introduction}

Recommending users with a set of preferred items is still an open problem~\cite{koren2009matrix,zheng2018spectral,he2017neural,man2017cross,elkahky2015multi,farseev2017cross}, especially when the dataset is very sparse. To remedy the data sparsity issue, broad-leraning based model~\cite{broad_learning} and cross-domain recommender system~\cite{hu2013personalized,man2017cross} are proposed where the information from other source domains can be transferred to the target domain. To transfer the knowledge from one domain to another, one can use the overlapping users~\cite{hu2013personalized,farseev2017cross,man2017cross,hu2018conet} in two ways: (1) the neighborhood information of common users stores the structure information of different domains with which we can do cross-domain recommendation~\cite{wang2018cross,farseev2017cross}; or (2) we can learn a mapping function~\cite{man2017cross,hu2013personalized} to project latent vectors learned in one domain into another, and thus the knowledge can be transferred. 

\begin{figure}[hbt!]
    \centering
    \includegraphics[height=2in]{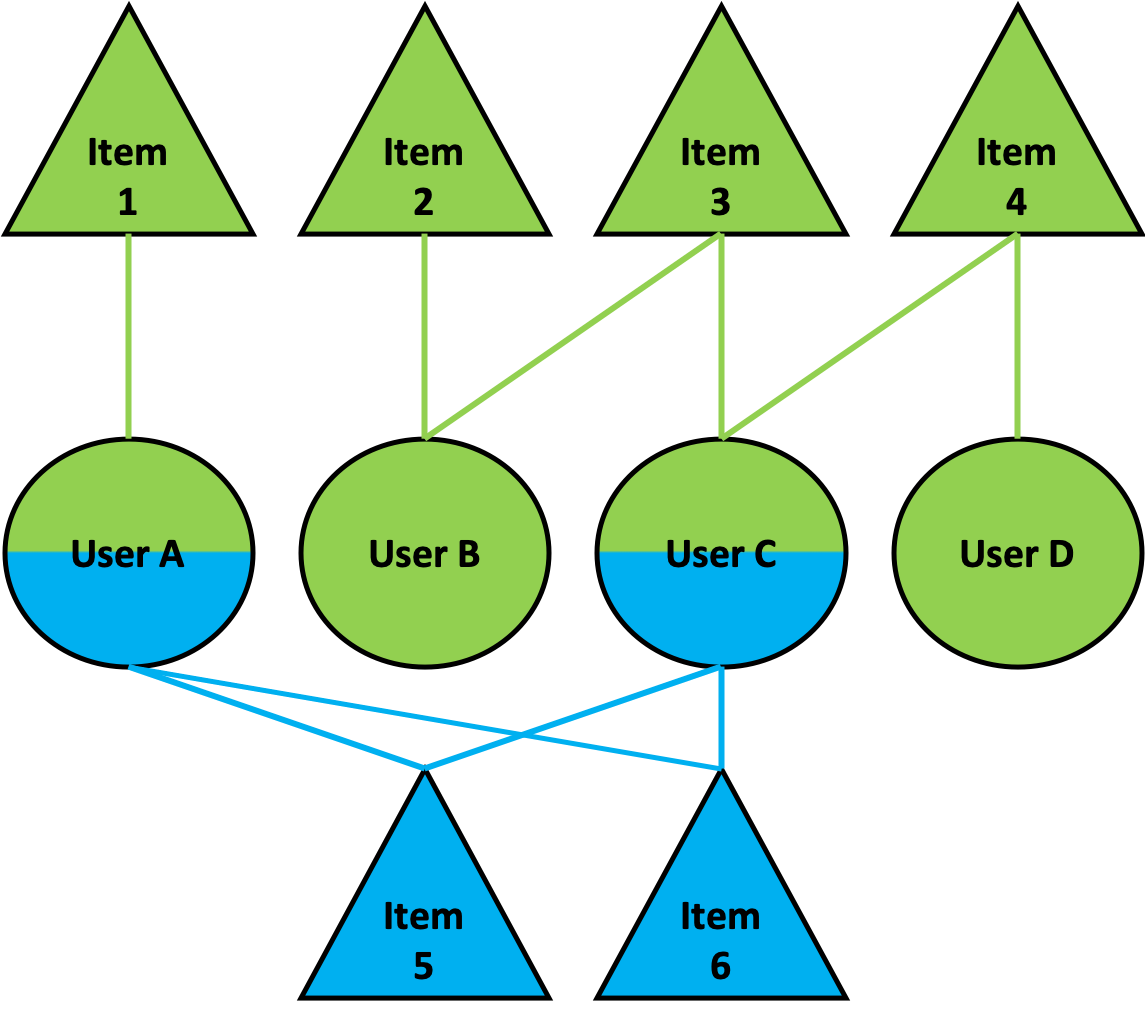}
    \caption{A toy example of high-order connectivity information in cross-domain recommendation. The upper/green part is the target domain and the below/blue part is the source domain.}
    \label{fig:toyexample}
\end{figure}

However, all existing methods ignore the \textit{high-order connectivity} information~\cite{6494675}.  High-order connectivity information consists of all the neighborhood information, the neighbors of all the neighbors, and so on by using the linkage information in the graph. The high-order connectivity information is explained in Figure~\ref{fig:toyexample} wherein the middle part user A and user C are the overlapping users, the upper/green part is the target domain, and the lower/blue part is the source domain. For example, in the target domain \textit{(only the upper part)}, user D has a connection with item $4$. Merely with the neighbor-based information~\cite{he2017neural,hu2013personalized,zhang2016collaborative}, item $1$ and item $2$ should be ranked similarly since the neighbor user (i.e. user C) of user D has no direct connections with them. However, with the high-order connectivity information, we argue that user D should prefer item $2$ more than item $1$ as there is a path from item 2 to user D \footnote{\textit{item 2--user B--item 3--user C--item 4--user D}}, while item $1$ is only connected with user A and apart from the others. Moreover, the preference ranking may be different if taking account of the source domain \textit{(considering both the upper and lower graphs)}. We can find two paths~\footnote{\textit{item 1--user A--item 5--user C--item 4--user D} and \textit{item 1--user A--item 6--user C--item 4--user D}} from item $1$ to user D compared with the single path from item $2$ to user D. Hence user D may prefer item $1$ more than item $2$ if the high-order connectivity information across domains is included. However, the high-order connectivity problem is not well studied yet in the cross-domain recommendation.


To capture the connectivity information in a graph, one can transform the graph into the frequency domain by applying the spectral graph theory~\cite{kipf2016semi,defferrard2016convolutional,6494675}. In spectral theory~\cite{6494675,chung1997spectral}, the spectrum of a graph extracts the comprehensive connectivity information of a graph with the graph Fourier transformer in terms of the eigenvectors of the graph laplacian~\cite{6494675}. Based on this, we can design the spectral convolutional network~\cite{zheng2018spectral,bruna2013spectral} whose convolutions are linear operators diagonalized by the Fourier basis. With the spectral convolutional network, nodes in a graph are represented as spectral vectors~\cite{kipf2016semi,zheng2018spectral}. When it comes to bipartite graphs, we can learn the spectral representations of users and items to capture the connectivity information. The spectral representation models the high-order non-linear interactions among users and items with multi-layer spectral convolutions. Hence, recalling the problem discussed before in Figure~\ref{fig:toyexample}, in the spectral domain, the item 1 will be closer to user D than item 2 as there exist more connections from item 1 to user D than those from item 2. 

\begin{figure}
    \centering
    \includegraphics[height=1.5in]{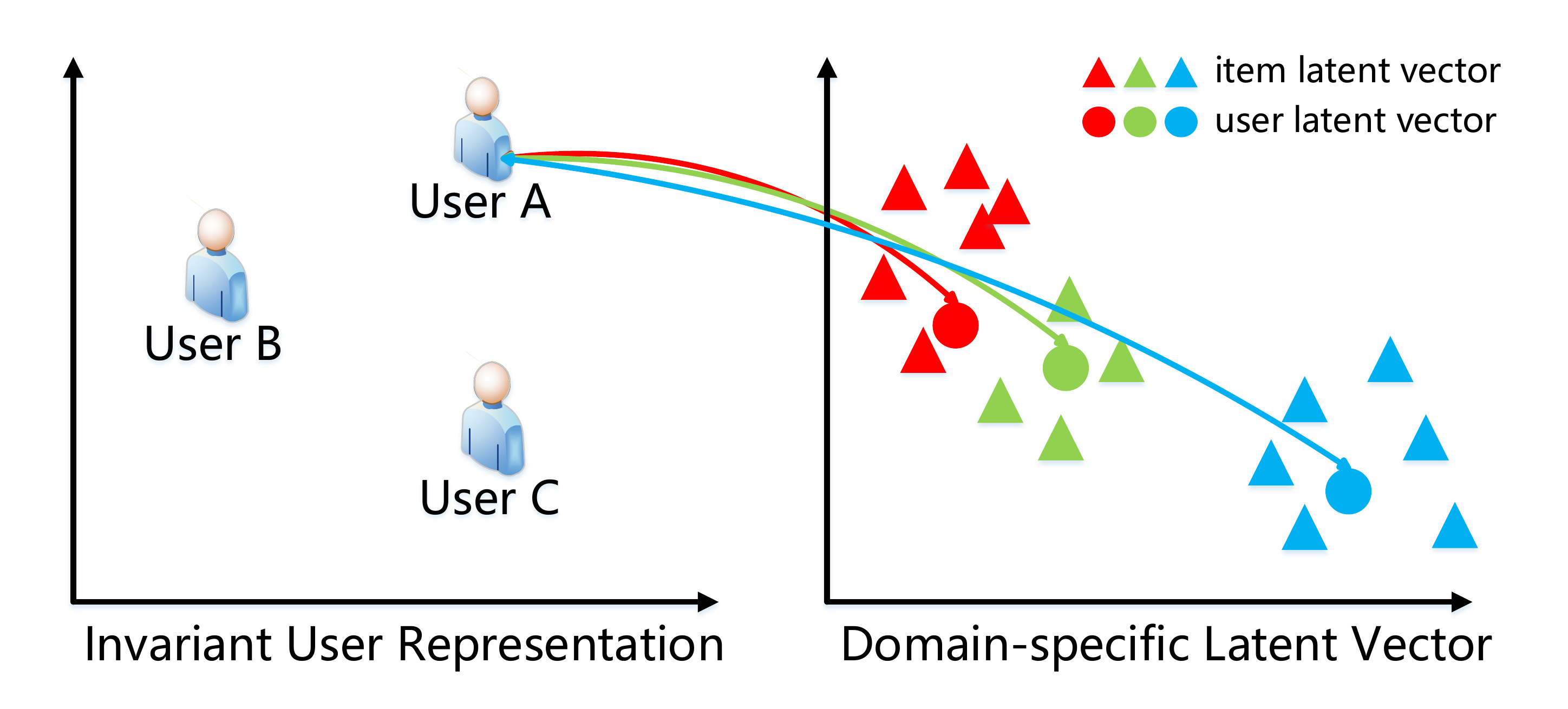}
    \vspace{-1em}
    \caption{Mapping users to different domains. Each user has domain invariant user representation on the left, which is projected as different domain-specific latent vectors over a specific domain on the right. Different colors on the right represent different domains. }
        \vspace{-1em}
    \label{fig:toy example}
\end{figure}

However, different domains may be incompatible with each other which is also called as \textit{domain-incompatibility} problem~\cite{shi2018easing} in the cross-domain recommendation. For instance, if the target domain is a \textit{Movie} domain where users are connected with the movie items, and the source domain is a \textit{Clothing} domain where users are connected with the clothing items, they will be incompatible with each other since the behavior of users varies a lot. The information from the source domain cannot be directly utilized in the target domain. Thus we need to propose some mapping methods~\cite{man2017cross,kazama2016cross,li2014matching} as a bridge for the information transferring. 

In this paper, unlike previous direct mapping methods~\cite{man2017cross,kazama2016cross,wang2018cross}, we view the latent vectors of a user in a specific domain as an interest projection from a \textit{domain-invariant representation}. We show an illustration of mapping the domain-invariant user representation to a domain-specific user latent vectors in Fig.~\ref{fig:toy example}. To learn transferable representations, we jointly learn the domain-invariant representation of users across different domains. The joint convolution can capture the high-order connectivity information across different domains and learn domain-invariant representations by keeping the spectral similarity of the overlapping users. Based on this, we design a Joint Spectral Convolutional Network (JSCN) to fuse the information from multiple domains. JSCN will simultaneously operate multi-layer spectral convolution on the graph from each domain. Then the extracted spectral features can be shared across different graphs with the domain-invariant representations. Since JSCN jointly learns the spectral representations on different graphs, the high-order comprehensive connectivity information can be shared across domains. And because of the domain-invariant representations of users, JSCN alleviates the domain-incompatibility problem. We summarize our main contributions as follows:
\begin{itemize}[leftmargin=*]

    \item \textbf{Transferable spectral representation:} To the best of our knowledge, it is the first work to study how to transfer the spectral representation of bipartite graphs, which captures the high-order non-linear interactions of user-item both within domain and across domains.
    
    \item \textbf{Joint spectral convolution on graphs:}
    In this paper, we design a joint spectral convolutional network for learning the representations of multiple graphs concurrently. The high-order comprehensive connectivity information can be shared across different graphs.


    \item \textbf{Domain adaptive module:} To deal with the domain incompatibility problem, we apply a novel domain adaptive module to jointly learn the domain-invariant spectral representations of users, with which we can implement the joint convolution on graphs and share information across different domains.
\end{itemize}

The rest of the paper is organized as follows. In Sec.~\ref{Related Work}, we review some previous works related to this paper. Then in Sec.~\ref{Preliminaries}, we introduce the definitions of the notations and concepts, as well as the problem. In Sec.~\ref{Proposed Model}, we present the proposed model and the formulation of the model. Finally, in Sec.~\ref{Experiment} we discuss the experiment before we draw a conclusion in Sec.~\ref{Conclusion}.

\section{Related Work}\label{Related Work}
In this section we give a brief review of two closely related areas: (1) deep learning based recommender system; and (2) cross-domain Recommendation.

\subsection{Deep learning based recommender system}
Since \cite{salakhutdinov2007restricted} introduces deep learning into recommender system (RS), \cite{zheng2016neural,he2017neural,hidasi2015session} propose deep neural network based RS to learn from either explicit or implicit data. To counter the sparsity problem, some scholars propose to utilize deep learning techniques to build a hybrid recommender system. \cite{van2013deep} and \cite{wang2014improving} introduce Convolutional Neural Networks (CNN) and Deep Belief Network (DBN) assist in representation learning for music data. These approaches above pre-train embeddings of users and items with matrix factorization and utilize deep models to fine-tune the learned item features based on item content.  
In \cite{elkahky2015multi}, a multi-view deep model is built to utilize item information from more than one domain. \cite{kim2016convolutional} integrates a CNN with PMF to analyze documents associated with items to predict users' future explicit ratings. \cite{zheng2017joint} leverages two parallel neural networks to jointly model latent factors of users and items. To incorporate visual signals into RS, \cite{he2016vbpr, liu2017deepstyle,mcauley2015image,he2016ups} propose CNN-based models to incorporate visual signals into RS. They make use of visual features extracted from product images using deep networks to enhance the performance of RS. \cite{zhang2016collaborative, zhang2017joint} investigates how to leverage the multi-view information to improve the quality of recommender systems. Due to the limited space, readers can refer to \cite{zhang2017deep} for more works on deep recommender systems.

\subsection{Cross-domain recommendation and broad learning}
Broad Learning~\cite{broad_learning} is a way to transfer the information from different domains, which focuses on fusing and mining multiple information sources of large volumes and diverse varieties. To solve the \textit{cold-start} problem in item recommendation, cross-domain recommendation is proposed by either learning shallow embedding with factorization machine~\cite{hu2013personalized,singh2008relational,loni2014cross,wang2018cross} or learning deep embedding with neural networks~\cite{man2017cross,misra2016cross,hu2018mtnet,hu2018conet,liu2019deep}. When learning shallow embedding, CMF~\cite{singh2008relational} jointly factorizes the user-item interaction matrices from different domains. In order to model the domain information explicitly, CDTF~\cite{hu2013personalized} and CDCF~\cite{loni2014cross} is designed where the former factorizes the \textit{user-item-domain} triadic relation and the later models the source domain information as the context information of users. When learning the deep embedding of users and items, CSN~\cite{misra2016cross} is introduced firstly in multi-task learning scenario, where a convolutional network with cross-stitch units can share the parameters across different domains. This idea is extended later by CoNet~\cite{hu2018conet} with cross connections across different networks where shared mapping matrices is introduced to transfer the knowledge. Additionally, EMCDR~\cite{man2017cross} transfers the knowledge across source and target domains with multi-layer perceptron. Our proposed JSCN model also jointly learns a deep embedding for both in-domain and cross-domain information.

\section{PRELIMINARIES and Definition}\label{Preliminaries}

In this section, the preliminaries and definitions are presented. At first, we formally define the user-item bipartite graph and the corresponding connectivity matrices. Then we define the bipartite graph domain as well as the source domain and target domain before we formulate our problem. The important notations used in this paper are summarized in Table~\ref{notation table}.

\begin{mydef}
    \textbf{(Bipartite Graph)}. A bipartite user-item graph $\mathcal{B}$ with $N$ vertices and $E$ edges for recommendation is defined as $\mathcal{B}=\{\mathcal{U},\mathcal{I},\mathcal{E}\}$, where $\mathcal{U}$ and $\mathcal{I}$ are two disjoint vertex sets, i.e. user set and item set, respectively. Every edge $e\in\mathcal{E}$ is in the form as $e = (u,i)$, denoting the interaction of a user $u\in \mathcal{U}$ with an item $i \in \mathcal{I}$, e.g. an item is viewed/purchased/liked by a user.
\end{mydef}



A bipartite graph describes the interactions among users and items, thus we can define an \textit{implicit feedback matrix}~\cite{rendle2009bpr,he2017neural} $\mathbf{R} \in \{0,1\}^{|\mathcal{U}| \times |\mathcal{I}|}$ for a given bipartite graph $\mathcal{B}$ as

\begin{equation}\label{feedback matrix}
\mathbf{R}_{r,j} = \left\{\begin{matrix}
1 & \text{if $(u_r, i_j)$ interaction is observed} \\
0 & \text{otherwise},
\end{matrix}\right.
\end{equation}
where $u_r$ and $i_j$ are the $r$-th user in the user set $\mathcal{U}$ and $j$-th item in the item set $\mathcal{I}$, respectively.

Given an implicit feedback matrix $\mathbf{R}$ of a bipartite graph $\mathcal{B}$, the corresponding \textit{adjacent matrix} $\mathbf{A}$ can be defined as
\begin{equation}\label{adjacent matrix}
\mathbf{A} = \begin{bmatrix}
0 & \mathbf{R} \\
\mathbf{R}^\top & 0
\end{bmatrix},
\end{equation}
where the adjacent matrix $\mathbf{A}$ is an $N \times N$ matrix and $N$ is the number of nodes in the bipartite graph, i.e., $N = |\mathcal{U}|+|\mathcal{I}|$.

With the adjacent matrix of a bipartite graph, a \textit{laplacian matrix} $\mathbf{L}$ of a bipartite graph can be calculated as
\begin{equation}
\mathbf{L} = \mathbf{I} - \mathbf{D}^{-1}\mathbf{A},
\end{equation}
where $\mathbf{I}$ is the identity matrix and $\mathbf{D}$ is a diagonal matrix where each entry on the diagonal denotes the sum of all the elements in the corresponding row of the adjacent matrix, i.e. $\mathbf{D}_{k,k}=\sum_{t}\mathbf{A}_{k,t}$.


In this paper, we focus on the cross-domain recommendation. Thus we would combine the information from a set of bipartite graphs and then recommend items to users. In each domain, we have a categorical mapping function $\Psi$ which projects the items into a specific category, e.g. \textit{Movies}, describing the type of the items in the domain. We assume all the items belongs to one domain and thus we have the definition of graph domain.


\begin{mydef}
    \textbf{(Bipartite graph domain)} A Bipartite graph domain is defined on a categorical mapping function $\Psi$ of items. Two bipartite graphs $\mathcal{B}_1$ and $\mathcal{B}_2$ are in different domains if and only if $\Psi(\mathcal{I}_1) \neq \Psi(\mathcal{I}_2)$.
\end{mydef}


The \textbf{source domain} bipartite graph is the source interaction bipartite graph of users and items, which provides auxiliary information for \textbf{target domain} bipartite graph where we would recommend items to users. We would integrate the information across the source domain and target domain, and make a recommendation in the target domain.


\begin{mydef}\label{problem definition}
    \textbf{(Problem Definition)}. Given a set of source domain bipartite graphs $\{\mathcal{B}^{s}_{1}, \mathcal{B}^{s}_{2},...,\mathcal{B}^{s}_{M}\}$  and a target domain graph $\mathcal{B}^{t}$,  we aim at recommending each user in $\mathcal{U}^t$ with a ranked list of items from $\mathcal{I}^t$ which have no existing interaction with that user in graph $\mathcal{B}^{t}$. The source domains share a set of common users with each other, and the shared users between pairwise source domains can be denoted as set $\{\widetilde{\mathcal{U}}^{s}_{12}, \widetilde{\mathcal{U}}^{s}_{13},..., \widetilde{\mathcal{U}}^{s}_{(M-1)M}\}$. Meanwhile, the target domain also shares a set of common users with each of the source domains, which denoted as set $\{\widetilde{\mathcal{U}}^{t}_{1}, \widetilde{\mathcal{U}}^{t}_{2},..., \widetilde{\mathcal{U}}^{t}_{M}\}$.
\end{mydef}


\begin{table}
    \centering
    \vspace{-1em}
    \caption{Important notations}\label{notation table}
    \vspace{-1em}
    \begin{tabular}{ll}
        \hline\hline
        Notation & Description    \\ \hline\hline
        $\mathcal{B},\mathcal{B}^s,\mathcal{B}^t$      & bipartite graph, source graph, target graph                               \\ 
        $\mathcal{U}$      & set of users                               \\ 
        $\mathcal{I}$      & set of items                               \\
        $u,i$       & user, item            \\
        $\widetilde{\mathcal{U}}$  &  set of common users     \\
        $\widetilde{u}$  &  common user     \\
        $\mho,\Lambda$      & eigenvectors, diag-matrix of eigenvalues\\
        $C$ & input dimension of feature vector\\
        $F$                 & spectral convolution parameter in each layer \\
        $\mathbf{V}^{u}$,$\mathbf{V}^{i}$      & user, item latent vectors \\
        $\mathbf{U}^s$,$\mathbf{U}^t$       & source, target invariant user representation \\
        $d$                 & dimension of spectral latent vectors \\
        $d'$      &  dimension of domain-invariant representation\\
        $\Phi_{\mathcal{B}}$ & domain related user mapping function \\
        $\Psi$   & categorical mapping function of items\\
    
        \hline\hline
    \end{tabular}
\end{table}

\section{Proposed Model}\label{Proposed Model}
\begin{figure*}
    \centering
    \includegraphics[height=2.5in]{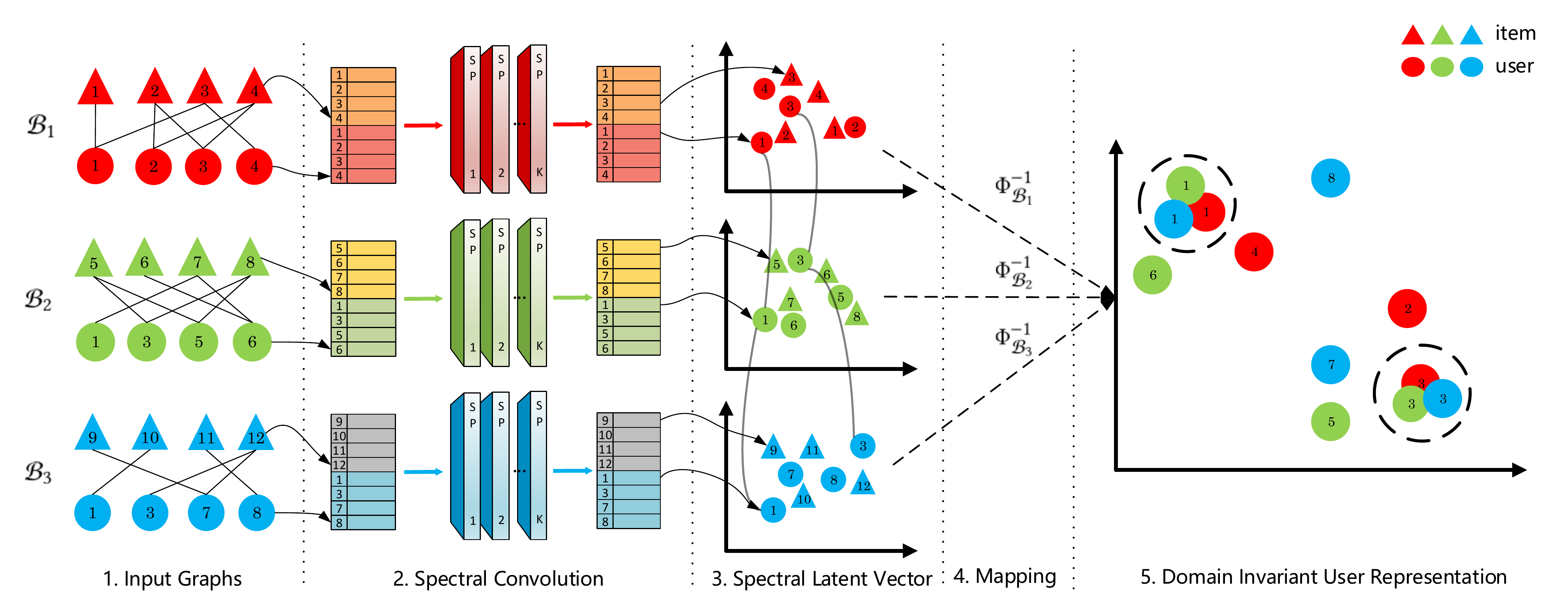}
    \vspace{-1em}
    \caption{The framework of training joint spectral convolution network (JSCN) model. At first, we randomly initialize the users (below part) and items (upper part) in the input graphs. Secondly, we learn spectral latent vectors of users and item with $K$-layer spectral convolution network (SP). Then we map the spectral latent vectors of users to domain invariant user representations with user mapping function $\Phi_{\mathcal{B}_i}^{-1}$. Finally, we minimize the distance of common users in domain invariant user representation space.}
    \vspace{-1em}
    \label{fig:framework}
\end{figure*}

In this section, we explain the spectral convolution network for collaborative filtering~\cite{zheng2018spectral} first before we introduce the domain invariant user representation. After that, we will present our proposed \textbf{J}oint \textbf{S}pectral \textbf{C}onvolution \textbf{N}etwork~(JSCN) for cross-domain recommendation. Finally, we will formulate the adaptive user mapping mechanism. The overall framework of our proposed model is given in Fig.~\ref{fig:framework}. We use triangles and squares denoting users and items, respectively. Different colors for users and items denote different domains. And the same numbers on squares represent common users in different domains.  

\subsection{Spectral Convolution on Graph}
Given a bipartite graph $\mathcal{B}=\{\mathcal{U},\mathcal{I},\mathcal{E}\}$, we would like to learn an embedding for each of the node, i.e. user or item, as illustrated in the first step in Fig.~\ref{fig:framework}. At first, users and items are represented as $C$-dimensional vectors, and all the user and item latent vectors can be grouped together and represented as matrices $\mathbf{X}^{u}$ and $\mathbf{X}^{i}$ respectively, where $\mathbf{X}^{u}\in \mathbb{R}^{|\mathcal{U}|\times C}$ and $\mathbf{X}^{i}\in \mathbb{R}^{|\mathcal{I}|\times C}$. With the graph structure information, the spectral convolutional operator $sp(\cdot)$ is defined~\cite{6494675,zheng2018spectral,defferrard2016convolutional} based on the eigenvectors $\mathbf{\mho}=\{\mathbf{\mu}_0, \mathbf{\mu}_1, \dots, \mathbf{\mu}_{N-1}\}$ and the corresonding eigenvalues $\Lambda = diag\{\lambda_0, \lambda_1, \dots, \lambda_{N-1}\}$ as

\begin{equation}\label{spectral operator}
    \begin{bmatrix}
\mathbf{X}^{u*}\\
\mathbf{X}^{i*}
\end{bmatrix}=sp(\begin{bmatrix}
\mathbf{X}^{u}\\
\mathbf{X}^{i}
\end{bmatrix}; \mho, \Lambda, \Theta)=\sigma \left((\mho\mho^\top+\mho\Lambda \mho^\top)\begin{bmatrix}
\mathbf{X}^{u}\\
\mathbf{X}^{i}
\end{bmatrix} \Theta \right ).
\end{equation}

In Eq.~(\ref{spectral operator}), the $\mho\mho^\top+\mho\Lambda \mho^\top$ term preserves the structure information of the bipartite graph, where $\Theta \in \mathbb{R}^{C\times F}$ is the convolutional filter to extract the spectral feature, and $\sigma(\cdot)$ denotes the logistic sigmoid function. It is the SP layer in the second step in Fig.~\ref{fig:framework}.

With multiple spectral convolutional operators on the original feature vectors $[\mathbf{X}^{u}_{0}, \mathbf{X}^{i}_{0}]^\top$, we construct a $K$-layer spectral convolutional network on bipartite graph as shown in Eq.~(\ref{multi-layer spectral}), with which we could learn the spectral representations of the nodes in the graph,

\begin{equation}\label{multi-layer spectral}
    \begin{bmatrix}
\mathbf{X}^{u}_{K}\\
\mathbf{X}^{i}_{K}
\end{bmatrix}=\underbrace{sp(\dots sp(}_{K}\begin{bmatrix}
\mathbf{X}^{u}_{0}\\
\mathbf{X}^{i}_{0}
\end{bmatrix}; \mho, \Lambda, \Theta_{0})\dots \mho, \Lambda, \Theta_{K-1}),
\end{equation}
where $\Theta_{0}\in \mathbb{R}^{C\times F}$ and $\Theta_{k}\in \mathbb{R}^{F\times F}$ ($k=1,2,\dots,K-1$).
After $K$-layer spectral convolutional operations, we represent the users and items as latent vectors $\mathbf{V}^{u}\in \mathbb{R}^{|\mathcal{U}|\times d}$ and $\mathbf{V}^{i}\in \mathbb{R}^{|\mathcal{I}|\times d}$ respectively by either concatenating the extracted spectral features vectors at each layer or using the spectral feature vectors at the last layer. It corresponds to the third step in Fig.~\ref{fig:framework}.

In terms of the loss function, we apply the \textit{BPR}-loss as suggested in~\cite{rendle2009bpr,zheng2018spectral} to compute the \textit{in-domain} loss, which models the in-domain user-item interactions,

\begin{equation}\label{in-domain loss}
\begin{split}
\mathcal{L}_{i} &= \sum_{(r,j,j')\in \mathcal{D}}-\ln \sigma \left( \mathbf{V}^{u}(u_r) \cdot \mathbf{V}^{i}(i_j)-\mathbf{V}^{u}(u_r) \cdot \mathbf{V}^{i}(i_{j'})\right).
\end{split}
\end{equation}
where the $\{(r,j,j')\}$ are the triples that sampled from user-item interaction records $\mathcal{D}$ in which $r$ denotes the index of a user, $j$ denotes the index of an item with which the user has interaction, and $j'$ denotes the index of an item with which the user has no interaction. And we apply dot product $\cdot$ of user vector and item vector. Unlike pair-wise learning process~\cite{koren2008factorization}, \textit{BPR}-loss maximizes the difference between $(r,j)$ and $(r,j')$ with the assumption that users prefer observed items $i_j$ over unobserved items $i_{j'}$.  We use $\mathbf{V}^{u}(u_r)$ denotes the user latent vector of user $u_r$, and $\mathbf{V}^{i}(i_j)$ and $\mathbf{V}^{i}(i_{j'})$ denote the item latent vector of item $i_j$ and item $i_{j'}$, respectively.

\subsection{Domain Invariant User Representation}~\label{domain invariant user section}
With the in-domain loss $\mathcal{L}_{i}$, we could learn both the user and item latent vectors from the multi-layer spectral convolutional network. Recall the problem definition in Def.~\ref{problem definition}, we have a set of source domain bipartite graphs $\{\mathcal{B}^{s}_{1}, \mathcal{B}^{s}_{2},...,\mathcal{B}^{s}_{M}\}$ and one target domain bipartite graph $\mathcal{B}^t$, and every domain has a set of overlapping users with each other. 

A user requires different aspects w.r.t. different domains that lead to different user latent vectors, but we prefer invariant user representation across different domains, and hence we define the domain invariant user representation as $\mathbf{U}\in \mathbb{R}^{|\mathcal{U}| \times d'}$, from which we generate the domain-specific latent vector $\mathbf{V}^u$ with corresponding domain-related user mapping function $\Phi_{\mathcal{B}}$ as $\mathbf{V}^u = \Phi_{\mathcal{B}}(\mathbf{U}).$


For example, $\mathcal{B}^s_m=\{\mathcal{U}^{s}_m,\mathcal{I}^{s}_m, \mathcal{E}^{s}_m\}$ has a set of common users with the target domain $\mathcal{B}^t=\{\mathcal{U}^{t},\mathcal{I}^{t}, \mathcal{E}^{t}\}$, which is denoted as $\widetilde{\mathcal{U}}^t_m$. With the in-domain loss,  we learn the domain specific user latent vectors individually for $\mathcal{B}^s_m$ and $\mathcal{B}^t$ as $\mathbf{V}^{us}_m$ and $\mathbf{V}^{ut}$ respectively. $\mathbf{V}^{us}_m$ is generated from the domain-independent user representations $\mathbf{U}^s_m$ by the corresponding domain-related user mapping function $\Phi_{\mathcal{B}^s_m}$. $\mathbf{V}^{ut}$ is generated from the domain-independent user representations $\mathbf{U}^t$ by the corresponding domain-related user mapping function  $\Phi_{\mathcal{B}^t}$. With the inverse function of the user mapping function $\Phi$, denoted as $\Phi^{-1}$, we can obtain the domain invariant user representation from the domain specific user latent vector as $\mathbf{U} = \Phi_{\mathcal{B}}^{-1}(\mathbf{V}^u)$, which is the fourth step in Fig.~\ref{fig:framework}.



Since we have the domain invariant user representations, each user in $\widetilde{\mathcal{U}}^t_m$ should be represented as a same representation both in $\mathbf{U}^s_m$ and $\mathbf{U}^t$. To make this constraint trainable, we construct the \textit{cross-domain} loss $\mathcal{L}_c$ as the $l_2$ distance of the domain invariant user representations  as:
\begin{equation}\label{cross domain loss}
\begin{split}
    \mathcal{L}_c &= \sum_{m=1}^{(M-1)}\sum_{n=m+1}^M\sum_{\widetilde{u}\in \widetilde{\mathcal{U}}_{mn}^{s}}\left\|\mathbf{U}_m^s(\widetilde{u}) - \mathbf{U}_n^s(\widetilde{u})\right\|_2 \\
       &+ \sum_{m=1}^M\sum_{\widetilde{u}\in \widetilde{\mathcal{U}}_{m}^{t}}\left\|\mathbf{U}_m^s(\widetilde{u}) - \mathbf{U}^t(\widetilde{u})\right\|_2,
\end{split}
\end{equation}


    

where $\widetilde{u} \in \widetilde{\mathcal{U}}_{mn}^{s}$ denotes the common users between source domains $m$ and $n$ as defined in Def.~(\ref{problem definition}). $\mathbf{U}_m^s(\widetilde{u})$, $\mathbf{U}_n^s(\widetilde{u})$ and $\mathbf{U}^t(\widetilde{u})$ denotes the domain invariant representation of the anchor user $\widetilde{u}$ w.r.t. the corresponding domain-independent user representations $\mathbf{U}^s_m$, $\mathbf{U}^s_n$ and $\mathbf{U}^t$, respectively.


\subsection{Joint Spectral Convolutional Network}

The cross-domain loss $\mathcal{L}_c$ combines the information across different domains with the domain invariant user representation of the common users. Even if a common user only exists in part of all the domains, the information can be shared across different domains, as the effect of collaborative filtering. But we cannot directly learn the domain invariant representation, and thus instead, we learn the user and item latent vectors with the in-domain loss. Then we apply the inverse function of the user mapping function to learn the domain-invariant user representations. And the cross-domain loss can be written as:

\begin{equation}\label{cross domain loss mapping}
\begin{split}
    \mathcal{L}_c &= \sum_{m=1}^{(M-1)}\sum_{n=m+1}^M\sum_{\widetilde{u}\in \widetilde{\mathcal{U}}_{mn}^{s}}\left\|\Phi_{\mathcal{B}_m^{s}}^{-1}\left(\mathbf{V}_m^{us}(\widetilde{u}) \right) - \Phi_{\mathcal{B}_n^{s}}^{-1}\left(\mathbf{V}_n^{us}(\widetilde{u}) \right)\right\|_2^2 \\
       &+ \sum_{m=1}^M\sum_{\widetilde{u}\in \widetilde{\mathcal{U}}_{m}^{t}}\left\|\Phi_{\mathcal{B}_m^{s}}^{-1}\left(\mathbf{V}_m^{us}(\widetilde{u}) \right) - \Phi_{\mathcal{B}^{t}}^{-1}\left(\mathbf{V}^{t}(\widetilde{u}) \right)\right\|_2^2,
\end{split}
\end{equation}


where $\mathbf{V}_m^{us}(\widetilde{u})$, $\mathbf{V}_n^{us}(\widetilde{u})$ and $\mathbf{V}^{ut}(\widetilde{u})$ denote the latent vector of the common user $\widetilde{u}$ w.r.t. the corresponding domain-specific user latent vectors $\mathbf{V}^s_m$, $\mathbf{V}^s_n$ and $\mathbf{V}^t$, respectively. We present this in the fifth step in Fig.~\ref{fig:framework}. Hence the joint spectral convolution model has the loss function as:

\begin{equation}
   \mathcal{L} = \sum_{m=1}^M\mathcal{L}_{im}^{s} + \mathcal{L}_{i}^t + \mathcal{L}_c + Reg,
\end{equation}
where $\mathcal{L}_{im}^s$ is the in-domain loss of the source domain $B_{m}^s$, $\mathcal{L}_i^t$ is the in-domain loss of the target domain $B^t$, and $Reg$ is the regularization term defined as:
\begin{equation}
    Reg = \epsilon \left( \sum_{m=1}^M \left\| \mathbf{V}^{us}_m \right \|_{2}^{2} + \left \| \mathbf{V}^{t} \right \|_{2}^{2} \right ),
\end{equation}
where $\epsilon$ is the regularization hyper-parameter. 

\subsection{Adaptive User Mapping Module}
As described in Sec.~\ref{domain invariant user section}, we can use the inverse function of the domain-related user mapping function to generate the domain-invariant user representation from the spectral user latent vector. We define this inverse function as the adaptive user mapping function, which can either be a linear mapping function or a neural network based non-linear function~\cite{he2017neural}. For simplicity, here we only present the linear mapping function, which leads to  
\begin{equation}\label{adaptive user mapping}
    \mathbf{U} = \mathbf{V}^u W_{\mathcal{B}},
\end{equation}
where the $W_{\mathcal{B}}$ is the domain adaptive matrix w.r.t. graph domain $\mathcal{B}$. This mapping function is a kind of structural regularization~\cite{zhou2011malsar} of different domains. It turns out the mapping can transfer the spectral information during the joint learning process. 

With this adaptive user mapping matrix, we can rewrite the cross-domain loss as:
\begin{equation}\label{adaptive cross domain loss}
\begin{split}
    \mathcal{L}_c &= \sum_{m=1}^{(M-1)}\sum_{n=m+1}^M\sum_{\widetilde{u}\in \widetilde{\mathcal{U}}_{mn}^{s}}\left\|\mathbf{V}_m^{us}(\widetilde{u})W_{\mathcal{B}_m^{s}} - \mathbf{V}_n^{us}(\widetilde{u})W_{\mathcal{B}_n^{s}} \right\|_2^2 \\
       &+ \sum_{m=1}^M\sum_{\widetilde{u}\in \widetilde{\mathcal{U}}_{m}^{t}}\left\|\mathbf{V}_m^{us}(\widetilde{u})W_{\mathcal{B}_m^{s}}  - \mathbf{V}^{t}(\widetilde{u})W_{\mathcal{B}^{t}} \right\|_2^2,
\end{split}
\end{equation}
where $W_{\mathcal{B}_m^{s}}$ and $W_{\mathcal{B}_n^{s}}$ are two adaptive user mapping matrix corresponding with the domain $\mathcal{B}_m^{s}$ and $\mathcal{B}_n^{s}$ respectively.

\subsection{Optimization and Prediction}
We follow the optimization approach in~\cite{hinton2012neural,zheng2018spectral} to learn the spectral latent vectors and domain invariant user mapping with RMSprop. The RMSprop is an adaptive version of gradient descent which controls the step size with respect to the absolute value of the gradient. It is done by scaling the updated value of each weight by a running average of its gradient norm.

For the prediction, we focus on improving the performance on the target domain. We use the spectral representation $\mathbf{V}^{u}$ and $ \mathbf{V}^{i}$ of users and items respectively in the target domain to make a recommendation. For a specific user $u_r$, we predict the user's preference over an item $i_j$ as $\mathbf{V}^{u}(u_r) \cdot \mathbf{V}^{i}(i_j)$, then we sort the preferences as the ranking list for recommendation.

\section{Experiment}\label{Experiment}
In this section, we introduce the dataset first. After that, we discuss the baselines that we compare in this paper. Then we give the experimental settings such as the evaluation metrics. Finally, we present the experiments in details. Through the experiment, we respond to the following research questions:
\begin{itemize}[leftmargin=*]
    \item \textbf{RQ1}: Does the source domain information help to improve the recommendation performance in target domain?
    \item \textbf{RQ2}: Will spectral feature be better in improving the cross-domain recommendation performance?
    \item \textbf{RQ3}: Can the adaptive user mapping help to transfer the information across different domains?
\end{itemize}{}

\subsection{Dataset}
In this paper, we use the Amazon rating dataset~\cite{he2016ups}, where we find the interactions of users and items. The rating data where a user rates an item scoring from $1$ to $5$ is from May 1996 - July 2014. The dataset consists of $24$ different domains, we present part of the statistics as in Table~\ref{tab:domain statistics simple}. The original dataset is the rating data, we follow the convention in~\cite{zheng2018spectral,he2017neural} to transform the data into implicit interactions. 


\begin{table}
\centering
\vspace{-1em}
\caption{Dataset statistics~\Romannum{1}}
\vspace{-1em}
\label{tab:domain statistics simple}
\begin{tabular}{l|l|l|l}
\hline
 Domain Name & \#~User & \#~Item  & \#~Rating \\ 
 \hline
  Movies and TV & $2,089$~k & $201$~k & $4,607$~k  \\
  Clothing, Shoes and Jewelry & $3,117$~k & $1,136$~k & $5,748$~k \\
 Apps for Android & $1,324$~k & $61$~k & $2,638$~k \\
 Amazon Instant Video * & $427$~k & $24$~k & $584$~k  \\
  \hline
\end{tabular}
\end{table}

Each domain shares a set of common users with other domains. In the experiment, we use the \textit{Amazon Instant Video} dataset as the target domain and the other $23$ domains as the source domains.
\begin{itemize}[leftmargin=*]
    \item \textbf{Target Domain}: Amazon Instant Video consists of $583,933$ ratings among $426,922$~users and $23,965$ videos originally. Following the convention~\cite{zheng2018spectral,he2017neural}, we ignore the users with less than $5$ interactions, and the final domain has $3,113$ users, $5,860$ items with $22,256$ ratings (connections), and the sparsity is $99.878\%$.

    \item \textbf{Source Domain}: We use the other datasets as the source domain and part of the statistics of the dataset are illustrated in Table~\ref{tab:domain statistics simple} and Table~\ref{tab:multi-source domain}. And in the experiment, we compare $23$ different source domains and illustrate their contributions to the target domain.  
    
\end{itemize}



\subsection{Baseline}
To answer the previous research questions, we compare our proposed model and methods with some state-of-the-art methods. The major task is defined in Def.~\ref{problem definition} which focuses on improving the recommendation performance in the target domain. And we categorize the baseline methods into two groups: (1)~\textbf{Single domain based methods.} To answer RQ1 we should compare our model with other models that are non-cross-domain, e.g., BPR~\cite{rendle2009bpr}, NCF~\cite{he2017neural},  and SpectralCF~\cite{zheng2018spectral}. (2)~\textbf{Cross-domain based methods.} For RQ2, we will investigate the capability of spectral feature in transferring the information across different domains, e.g., CMF~\cite{singh2008relational}, CDCF~\cite{loni2014cross}, CoNet~\cite{hu2018conet} and our proposed model JSCN. For RQ3, we compare the different version of our proposed model to study the function of the adaptive user mapping. We introduce these methods as followings:
\begin{itemize}[leftmargin=*]

    \item \textbf{BPR}~\cite{rendle2009bpr}: BPR is a \textbf{B}ayesian \textbf{P}ersonalized \textbf{R}anking based Matrix Factorization method, which introduces a pair-wise loss into the Matrix Factorization to be optimized for ranking~\cite{gantner2011mymedialite}.
    
    \item \textbf{NCF}~\cite{he2017neural}: \textbf{N}eural \textbf{C}ollaborative \textbf{F}iltering applies neural architecture replacing the inner product of latent factors. Thus it can model the non-linear interaction of items and users. 
    
    \item \textbf{SpectralCF}~\cite{zheng2018spectral}: \textbf{Spectral} \textbf{C}ollaborative \textbf{F}iltering is the SOTA work to learn the spectral feature of users and items, which is based on the BPR pair-wise loss.
    
    \item \textbf{CMF}~\cite{singh2008relational}: \textbf{C}ollective \textbf{M}atrix \textbf{F}actorization is a matrix factorization based cross domain rating prediction model. In this paper, we change the rating to 0/1 w.r.t. the implicit interaction of users and items.
    
    
    \item \textbf{CDCF}~\cite{loni2014cross}:  \textbf{C}ross-\textbf{D}omain \textbf{C}ollaborative \textbf{F}iltering method model the user-item interaction as the context feature for the factorization machine. With arbitrary source domain, CDCF can treat them as input feature of users, and learn the latent vectors for both users and items. 
    
    \item 
    \textbf{CoNet}~\cite{hu2018conet}: It is the SOTA deep learning method to learn a shared cross-domain mapping matrix such that the information can be transferred. CoNet enables dual knowledge transferring across domains by introducing cross connections from one base network to another and vice versa. We implement the model with the code published by the author~\footnote{\url{http://home.cse.ust.hk/~ghuac/conet-code_only-hu-cikm18-20181115.zip}}.
    
    \item \textbf{JSCN-$\alpha$}: \textbf{J}oint \textbf{S}pectral \textbf{C}onvolution \textbf{N}etwork is our proposed model to learn a cross-domain recommender system. It is based on graph convolutional network to transfer the spectral feature of users across different domains. This model is a simple version without the adaptive user mapping, only enforcing the spectral vector in different domains to be similar.
    
    \item \textbf{JSCN-$\beta$}: It is the complete version of our proposed model, which includes adaptive user mapping.
\end{itemize}

\subsection{Experimental Setting}
Different from the rating score prediction task, the interaction prediction models in this paper should predict items that are interacted with users in the top ranking list. Thus in the experiment, we utilize the \textit{Recall@K} and \textit{MAP@K} to evaluate the performance of models. We usually have thousands of valid items in a given domain, we use $\mathbf{K}=\{20,40,60,80,100\}$ to present the performance of models. 

\begin{figure}[hbt!]
\centering
\begin{subfigure}{0.23\textwidth}
    \centering
    \includegraphics[width=\textwidth]{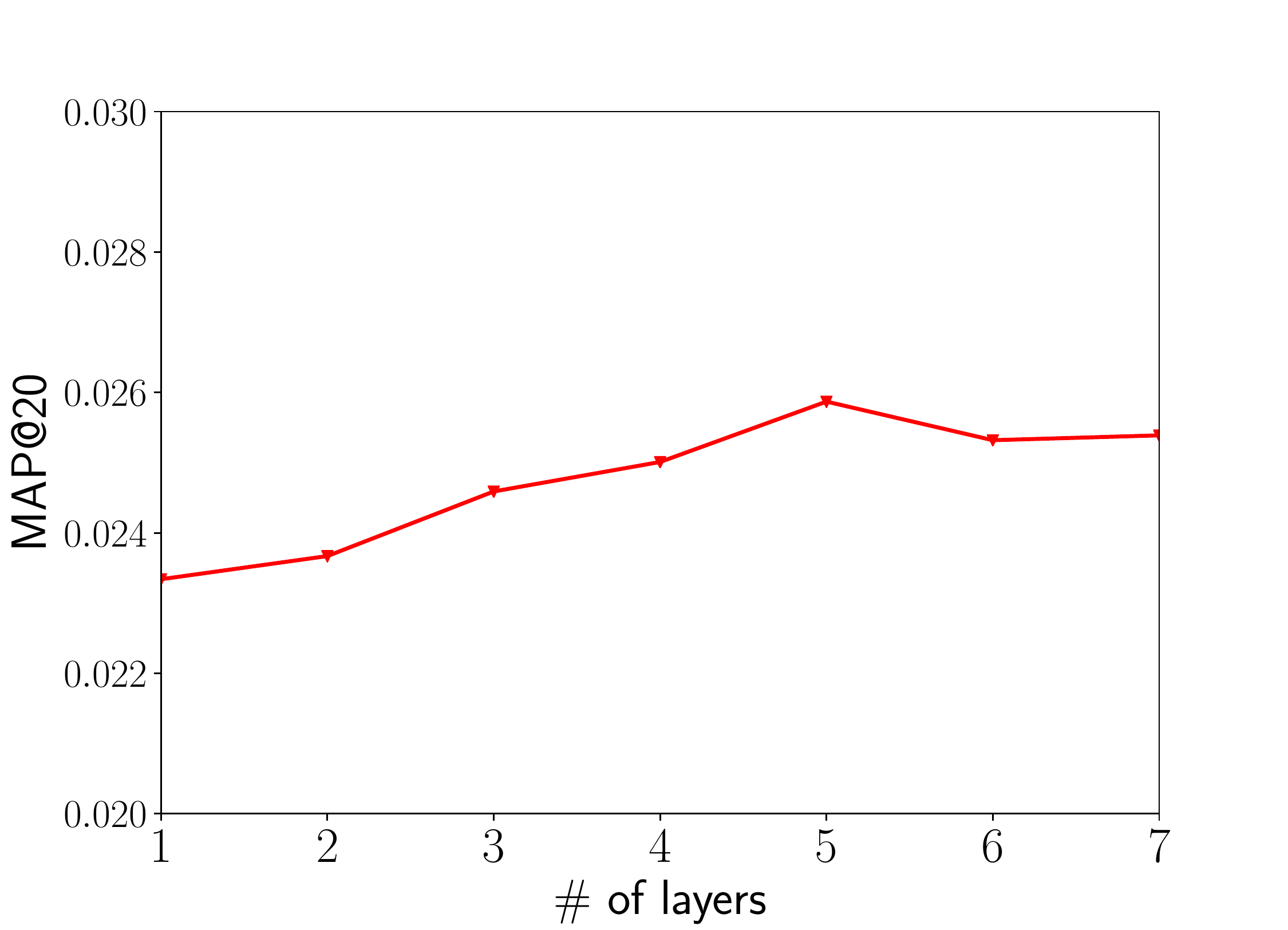}
    \label{fig:MAP_AfA_val}
\end{subfigure}
\begin{subfigure}{0.23\textwidth}
    \centering
    \includegraphics[width=\textwidth]{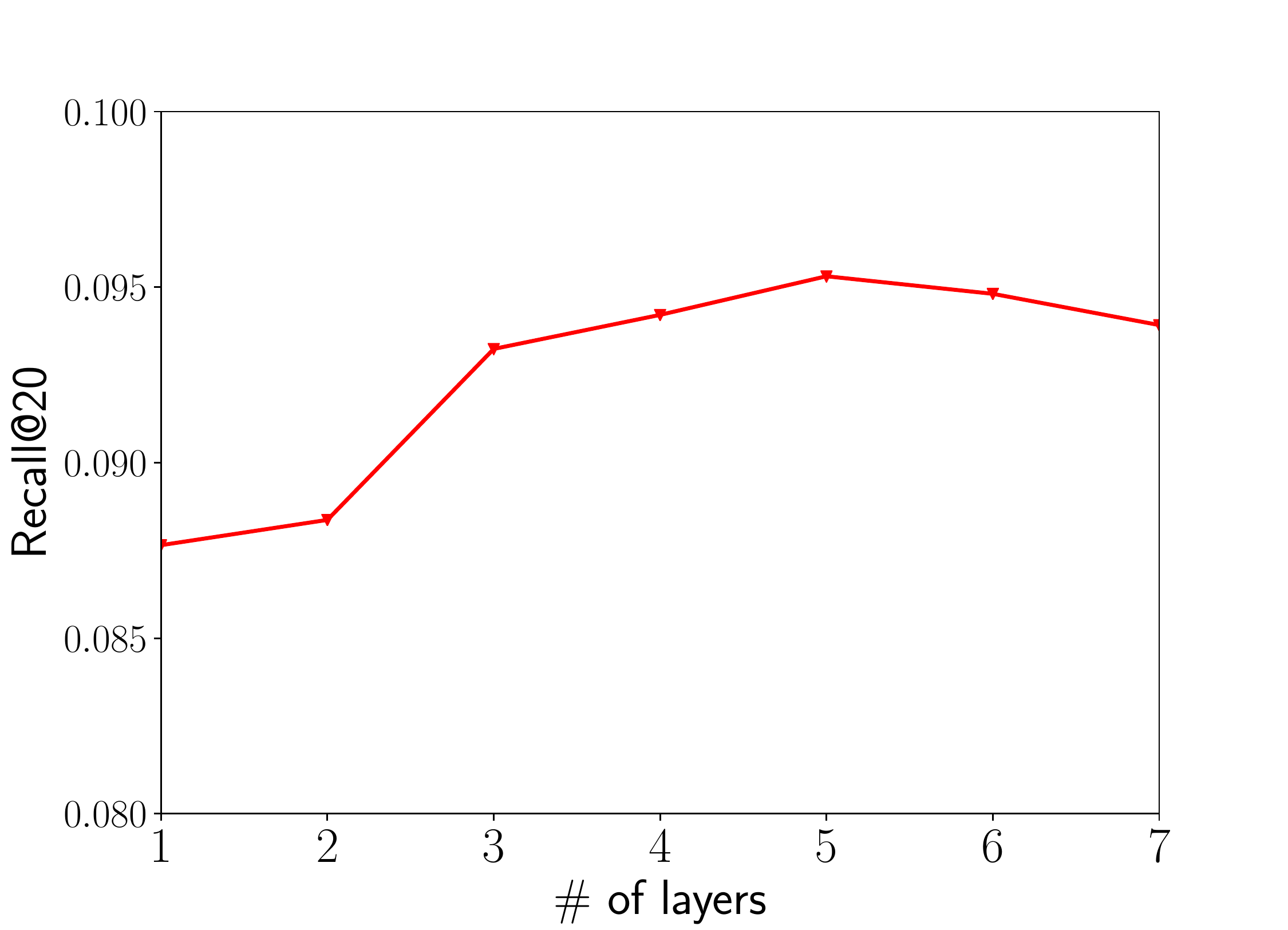}
    \label{fig:Recall_AfA_val}
\end{subfigure}
\caption{Validation performance of JSCN-$\beta$ for the hyper-parameter the number of convolutional layers w.r.t. \textbf{MAP@20} and \textbf{Recall@20} on target domain.}
\vspace{-1em}
\label{fig:validation}
\end{figure}

For the baseline methods, we select the dimension of latent vectors from ${8,16,32,64,128}$ for BPR and SpectralCF. And we follow the suggestion in original papers for NCF to train 3-layer MLP. We implement the CMF model by using the 0-1 interaction matrix. For CDCF, the dimension is set to $32$ which is the same as all the cross-domain based model. For our proposed model, there are some hyperparameters requiring tuning. To reduce the complexity of the proposed model, we would let the dimension of invariant user representation equal to the dimension of the spectral latent vector, i.e., $d'=d$. And we set the convolutional dimension parameter $F=C=32$. The number of filters is important to the performance of the model. And with the validation on different source domain datasets, we find when the number of filters $K=5$, the performance of JSCN is the best for most of the source domains. We present the validation on JSCN-$\beta$ with source domain as \textit{Apps for Android} in Figure~\ref{fig:validation}. And we use the linear mapping for domain adaptive part as suggested in Sec.~\ref{sec:domain adaptive}. For the training process, we set the learning rate as $0.001$ and the regularization weight $\epsilon$ as $0.001$.

\subsection{Cross-domain Comparison}
To answer \textbf{RQ1}, in this experiment part, we would compare the single domain based methods with the cross-domain based models on the same domain. The target domain is the \textit{Amazon Instant Video} dataset. And to answer \textbf{RQ2}, we would use the same source domain to compare different cross-domain based methods. To answer the \textbf{RQ3}, we would compare the performance of different versions of JSCN, i.e., JSCN-$\alpha$ and JSCN-$\beta$. In this section, we would use three different source domain datasets to improve the recommendation performance, which are \textit{Movies and TV}, \textit{Clothing, Shoes and Jewelry} and \textit{Apps for Android}. We analyze the results in details.

\begin{figure*}
\begin{subfigure}{.33\textwidth}
    \centering
    \includegraphics[width=\textwidth]{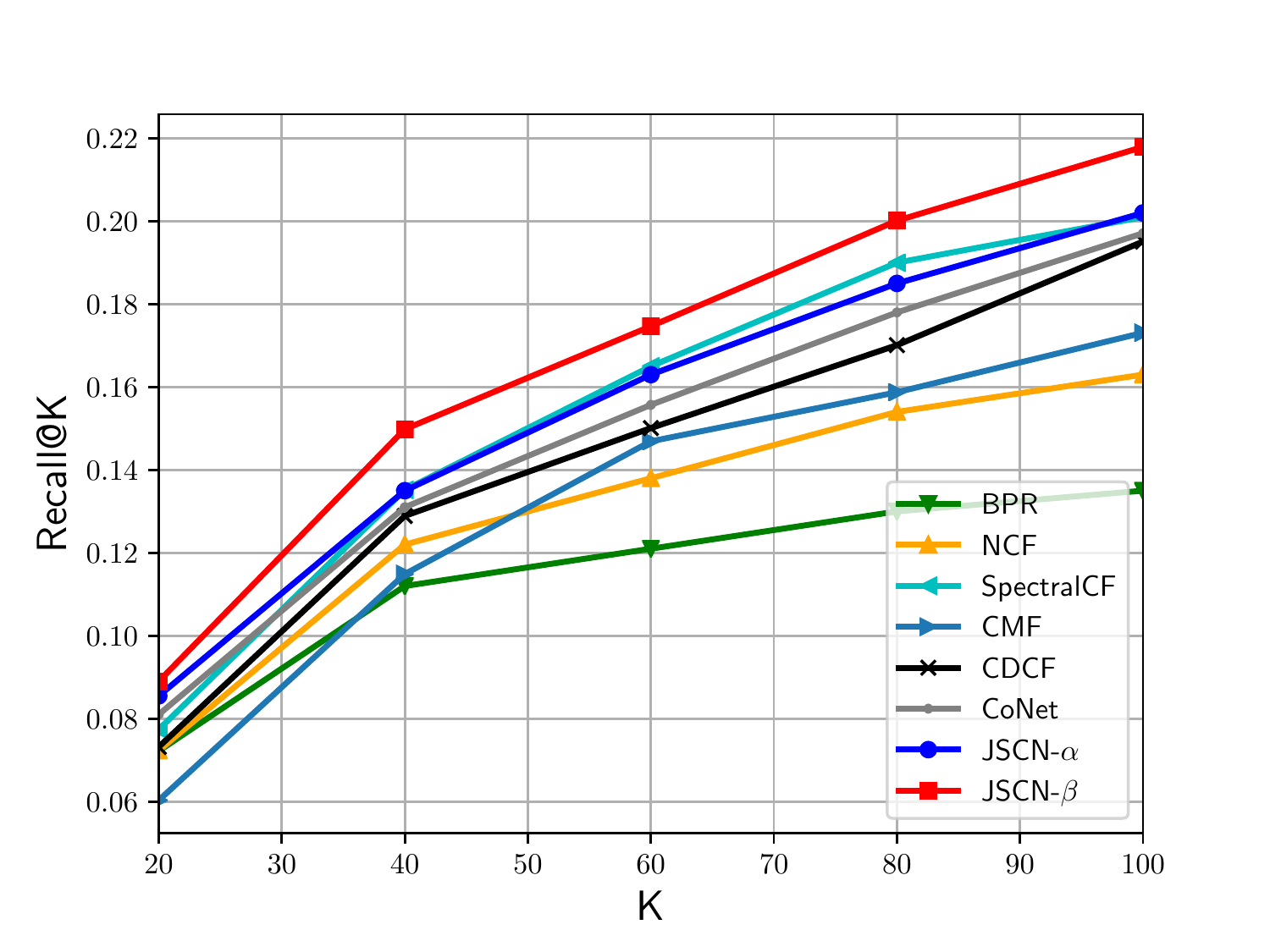}
    \caption{\textbf{Movies and TV}}
    \label{fig:recall_MTV}
\end{subfigure}
\begin{subfigure}{.33\textwidth}
    \centering
    \includegraphics[width=\textwidth]{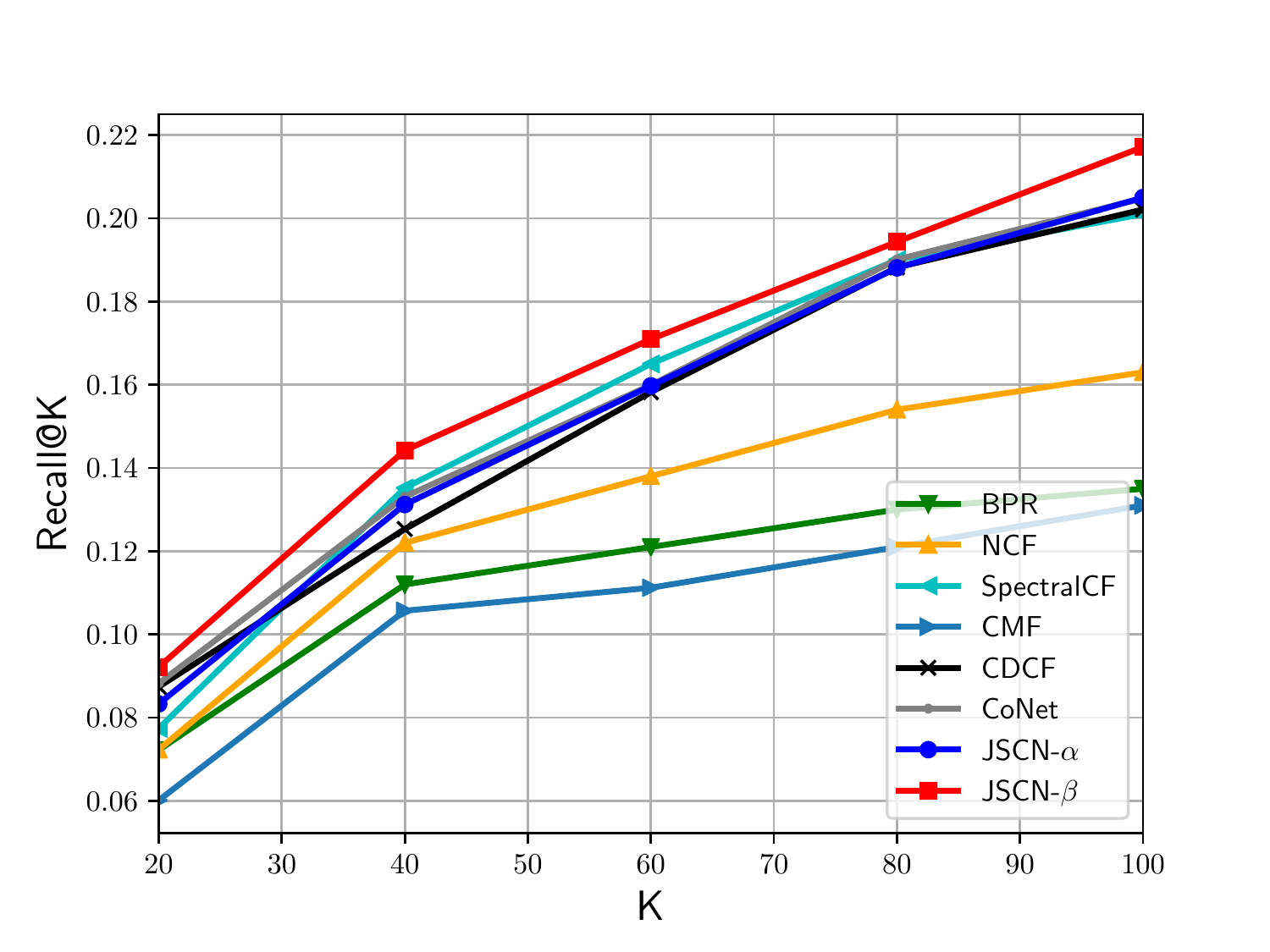}
    \caption{\textbf{Clothing, Shoes and Jewelry}}
    \label{fig:recall_CSJ}
\end{subfigure}
\begin{subfigure}{.33\textwidth}
    \centering
    \includegraphics[width=\textwidth]{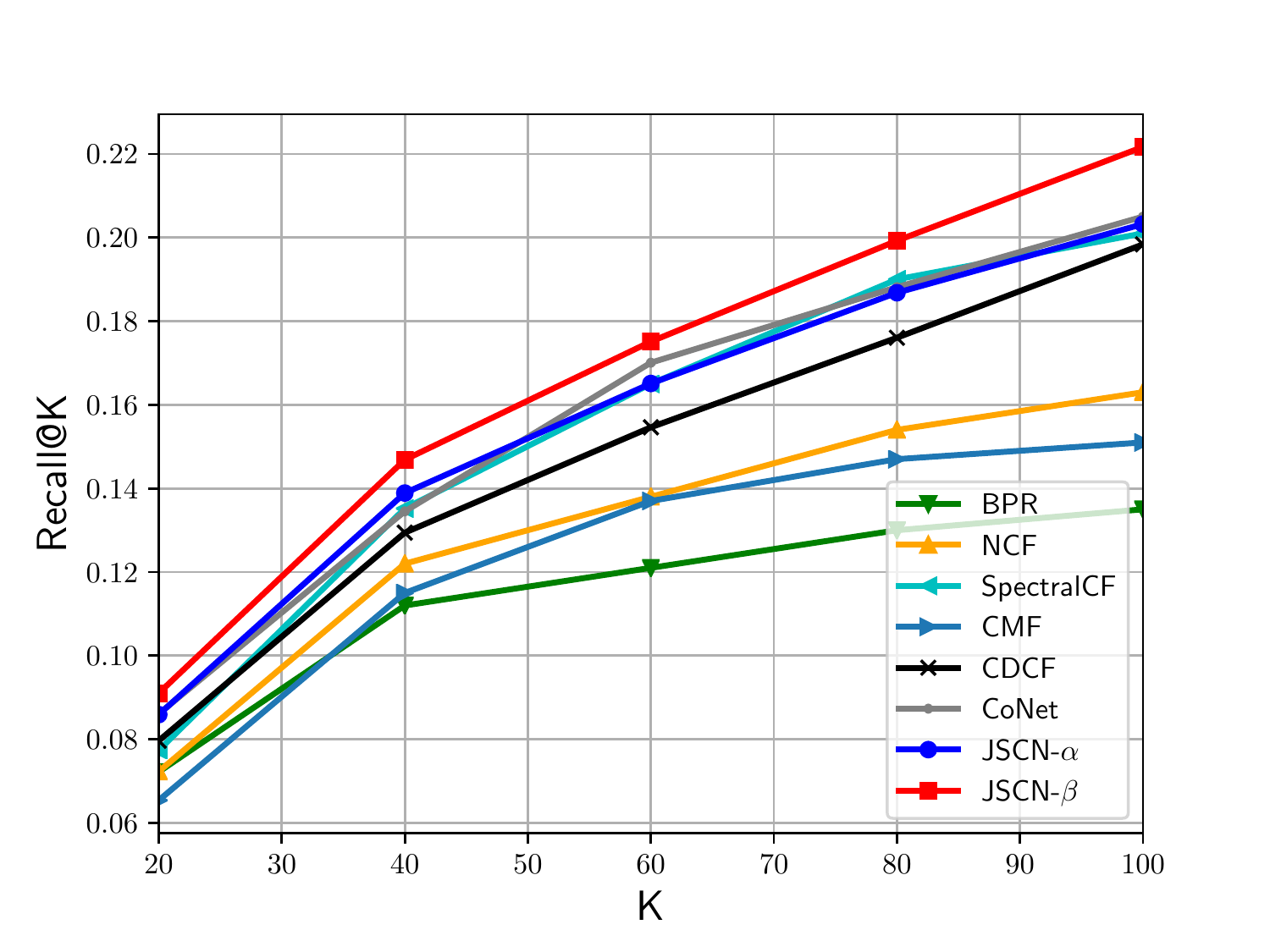}
    \caption{\textbf{Apps for Android}}
    \label{fig:recall_app}
\end{subfigure}
\vspace{-1em}
\caption{Performance comparison w.r.t. \textbf{Recall@K} on target domain \textit{Amazon Instant Video}, and with source domain \textit{Movies and TV},\textit{Clothing, shoes and Jewelry} and \textit{Apps for Android} respectively.}
\vspace{-1em}
\label{fig:recall_all}
\end{figure*}

\begin{figure*}
\begin{subfigure}{.33\textwidth}
    \centering
    \includegraphics[width=\textwidth]{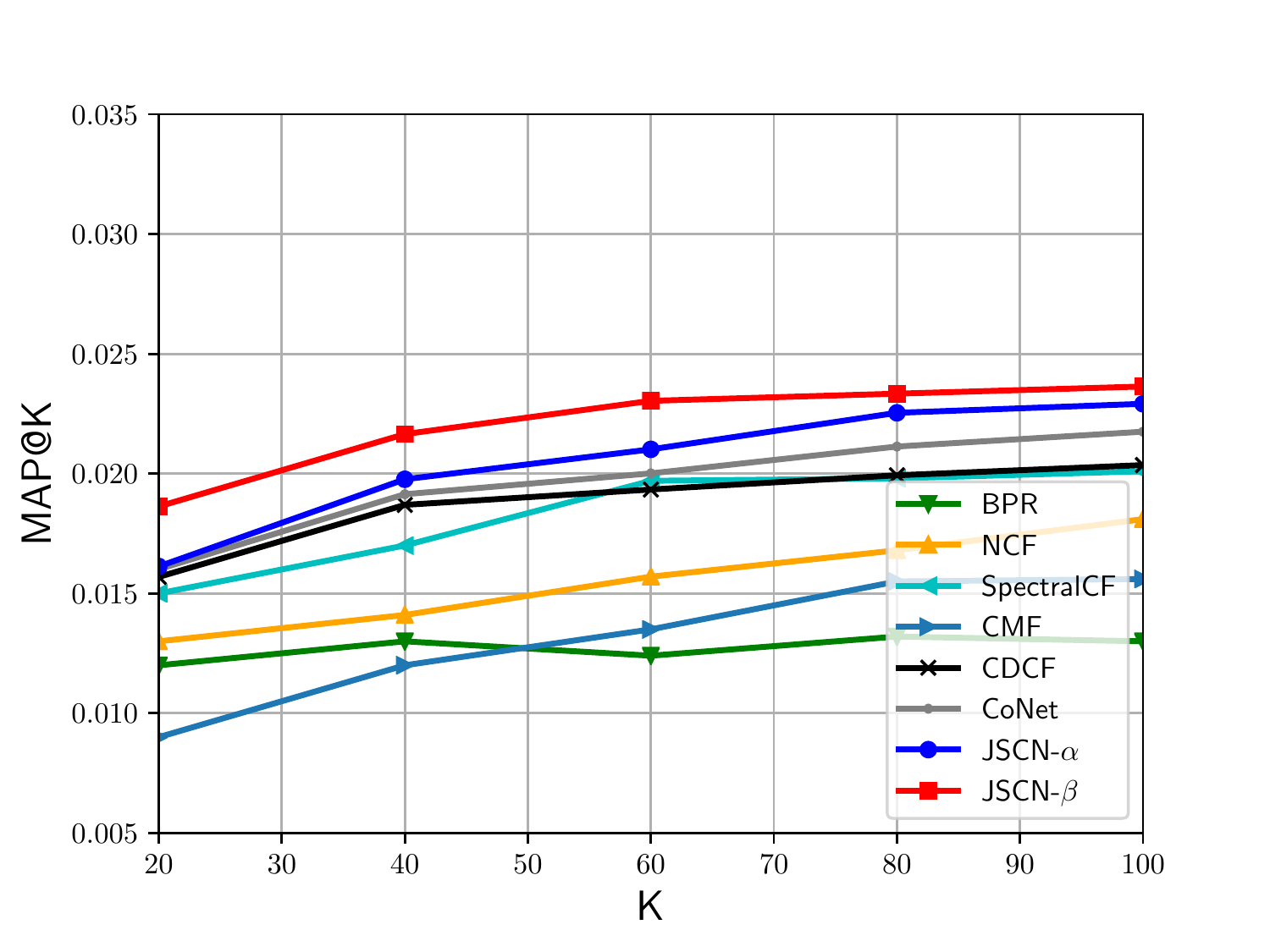}
    \caption{\textbf{Movies and TV}}
    \label{fig:MAP_MTV}
\end{subfigure}
\begin{subfigure}{.33\textwidth}
    \centering
    \includegraphics[width=\textwidth]{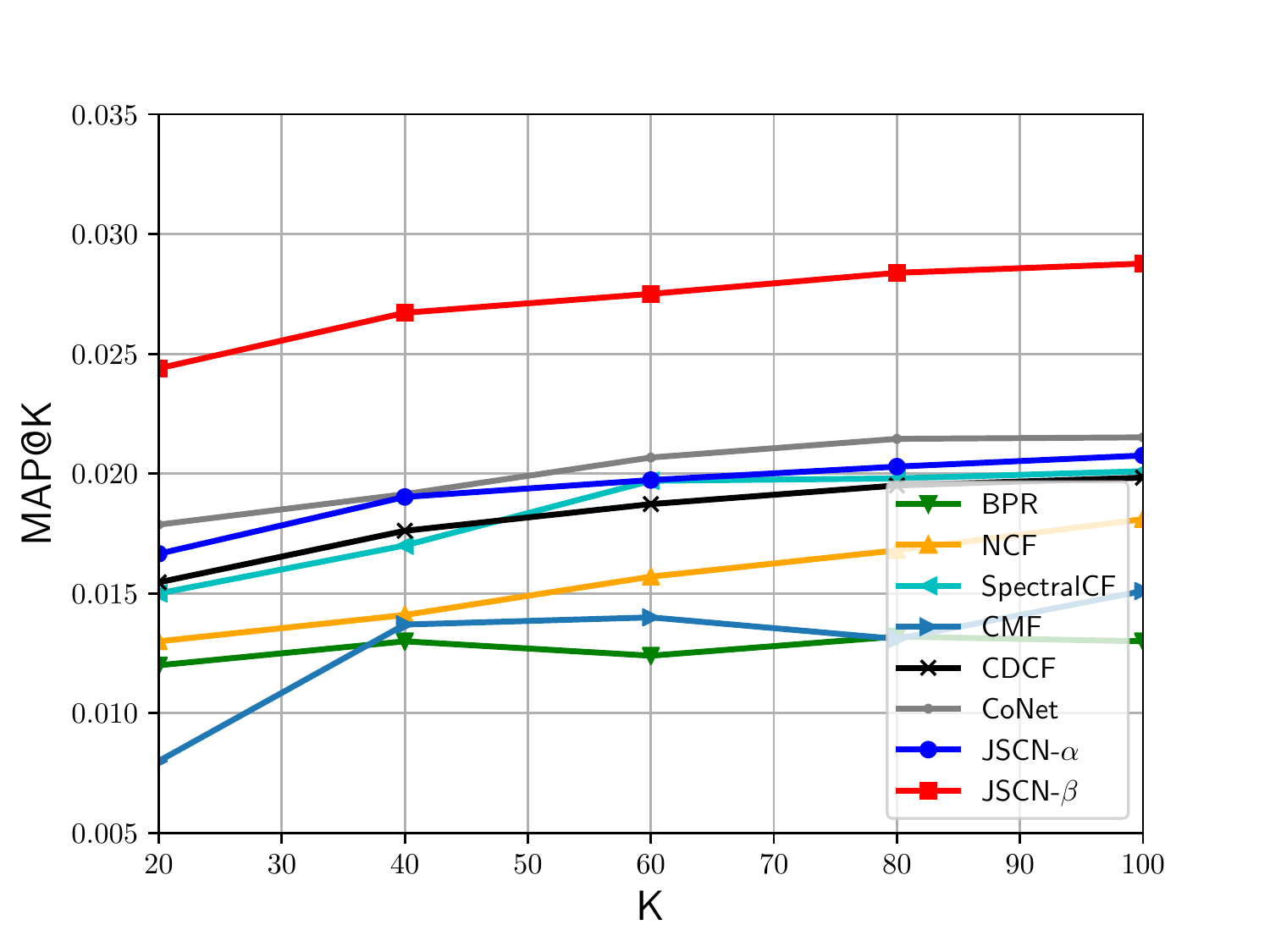}
    \caption{\textbf{Clothing, Shoes and Jewelry}}
    \label{fig:MAP_CSJ}
\end{subfigure}
\begin{subfigure}{.33\textwidth}
    \centering
    \includegraphics[width=\textwidth]{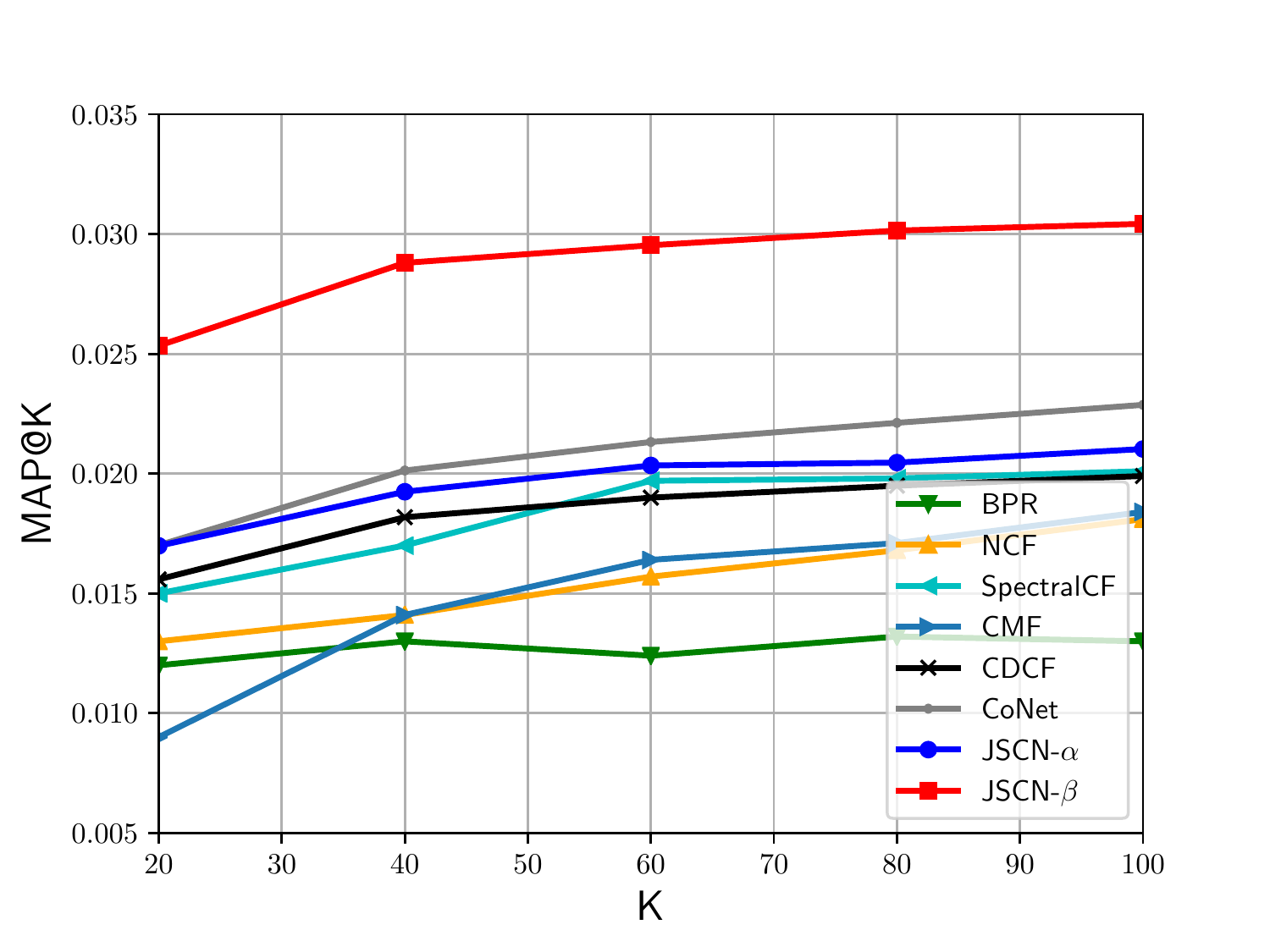}
    \caption{\textbf{Apps for Android}}
    \label{fig:MAP_app}
\end{subfigure}
\vspace{-1em}
\caption{Performance comparison w.r.t. \textbf{MAP@K} on target domain \textit{Amazon Instant Video}, and with source domain \textit{Movies and TV},\textit{Clothing, Shoes and Jewelry} and \textit{Apps for Android} respectively.}
\vspace{-1em}
\label{fig:MAP_all}
\end{figure*}

In Fig.~\ref{fig:recall_all}, we present the performance of different models on the target domain w.r.t. \textit{Recall@K}. And in Fig.~\ref{fig:MAP_all}, we show the performance w.r.t. \textit{MAP@K}.
For the cross-domain based models, JSCN-$\beta$ performs the best compared to all the other methods. JSCN-$\beta$ improves the performance of SpectralCF by $10.2\%$ on recall on average, and $38.3\%$ on MAP on average, which answers that cross-domain information can improve the performance. CMF cannot achieve a good performance compared to the other cross-domain based models. Among all the single domain based models, according to the result in~\cite{zheng2018spectral} and our results, SpectralCF is the best model compared to NCF and BPR as it can not only model the positive and negative interactions of user-item but also, with the graph convolution, model the interaction in a high-order non-linear way. From the result, some cross-domain based models cannot always surpass the single domain based models. 

CDCF, CoNet, JSCN-$\alpha$, and JSCN-$\beta$ can all well transfer the information across different domains. But since CDCF and CoNet has no spectral convolutional architecture, it cannot capture the high-order interactions of user-item. From our results, SpectralCF can achieve comparable performance with CDCF and CoNet even without source domain information. This suggests that we should apply spectral convolution to transfer the information across different domains. CoNet can transfer the information that learned from the neural networks and shared across different networks. But it cannot capture the high-order information across domain. JSCN-$\beta$ beats the performance of CoNet by $13.2\%$ on recall in average and $35.2\%$ on MAP in average, which answers that the spectral representation generated by JSCN can improve the performance in cross-domain recommendation.

The users in source domain \textit{Movies and TV} should request similar aspects of items with the users in target domain \textit{Amazon Instant Video} as the items are similar. Thus it is straightforward to transfer the information across these two compatible domains. The result is illustrated in Fig.~\ref{fig:recall_MTV} and Fig.~\ref{fig:MAP_MTV}. The performance of JSCN-$\beta$ and JSCN-$\alpha$ are relatively close. However, the source domain \textit{Clothing, Shoes and Jewelry} is incompatible with the target domain. From the result in Fig~\ref{fig:recall_CSJ} and Fig~\ref{fig:MAP_CSJ}, we can find both JSCN-$\alpha$ and CDCF cannot improve the performance compared to SpectralCF. But JSCN-$\beta$ learns the domain-invariant user representation which can transfer the information even the domain is incompatible. As a result, the adaptive user mapping in JSCN-$\beta$ is important to transfer the information across different domains even if the domains are incompatible. JSCN-$\beta$ beats the performance of JSCN-$\alpha$ by \textbf{9.2\%} on recall in average and \textbf{36.4\%} on MAP in average, which answers that the adaptive user mapping can solve the domain incompatible problem thus improve the performance in cross-domain recommendation.

\subsection{Comparison with Different Source Domains}\label{all source domains}

\begin{figure}
    \centering
    \includegraphics[width=0.45\textwidth]{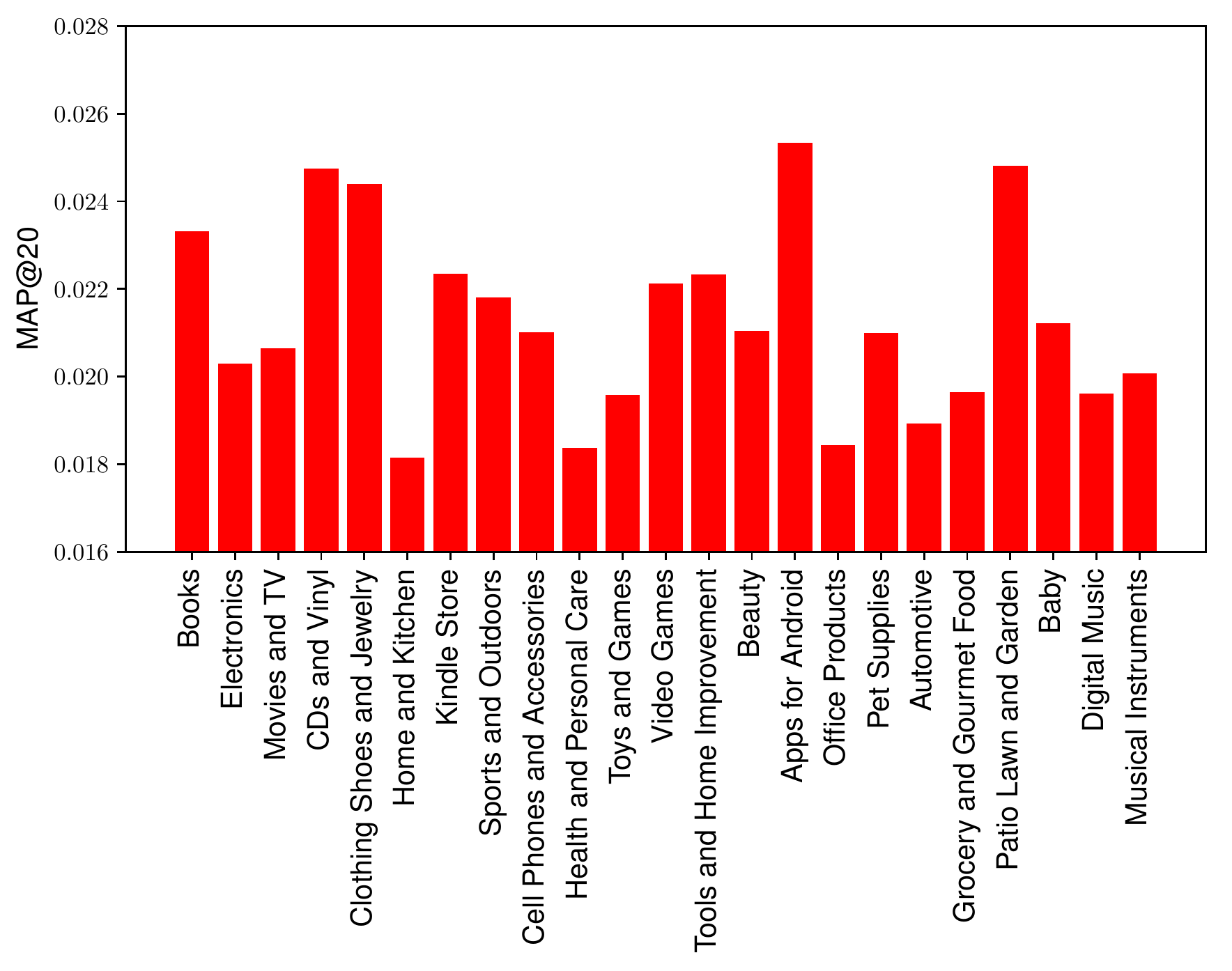}
    \vspace{-1em}
    \caption{Performance of JSCN-$\beta$ w.r.t. MAP@20 of the cross-domain recommendation on target domain.}
    \vspace{-1em}
    \label{fig:MAP_all_source}
\end{figure}

In this section, we report the cross-domain recommendation results of JSCN-$\beta$ on the target domain with different source domains w.r.t. \textbf{MAP@20} in Fig.~\ref{fig:MAP_all_source}. Since the recall performance varies little for different source domains, and due to the space limitation of the paper, we choose not to show the results of recall.

The best result is from source domain \textit{Apps for Android}. And we can find that even if some of the source domains are incompatible with the target domain \textit{Amazon Instant Video}, e.g. \textit{Clothing, Shoes and Jewelry}, the cross-domain recommendation performs well. Even if some of the source domains e.g. \textit{Home and Kitchen}, \textit{Health and Personal Care}, and \textit{Office Products}, perform not that well compared with other source domains, they still improve the performance of SpectralCF by $12.5\%$, $13.9\%$ and $14.3\%$ respectively, which suggests the benefits of source domain information and the effectiveness of our proposed model. 

\subsection{Multi-Source JSCN}

\begin{table}
\centering
\caption{Dataset Statistics \Romannum{2}}\label{tab:multi-source domain}
\vspace{-1em}
\begin{tabular}{c|l|l|l|l}
\hline
 label & Domain Name & \#~User & \#~Item  & \#~Ratings \\ 
 \hline
 ${1}$ & Home and Kitchen & $2,512$~k & $410$~k & $4,254$~k  \\
 2 & Health \& Personal Care & $1,851$~k & $252$~k & $2,982$~k \\
 3 & Office Products & $909$~k & $130$~k & $1,243$~k \\
  \hline
\end{tabular}
\end{table}

From the results in Sec.~\ref{all source domains}, we notice that our model performs differently given different source domain. Some source domains cannot provide enough information and hence the cross-domain recommendation results are not that good compared to the other source domains. The JSCN model can combine the information from $M$ source domains and share the information together to improve the performance on the target domain. In this section, we conduct the experiment on training JSCN models on multiple source domains.

\begin{figure}
    \centering
    \includegraphics[width=0.45\textwidth]{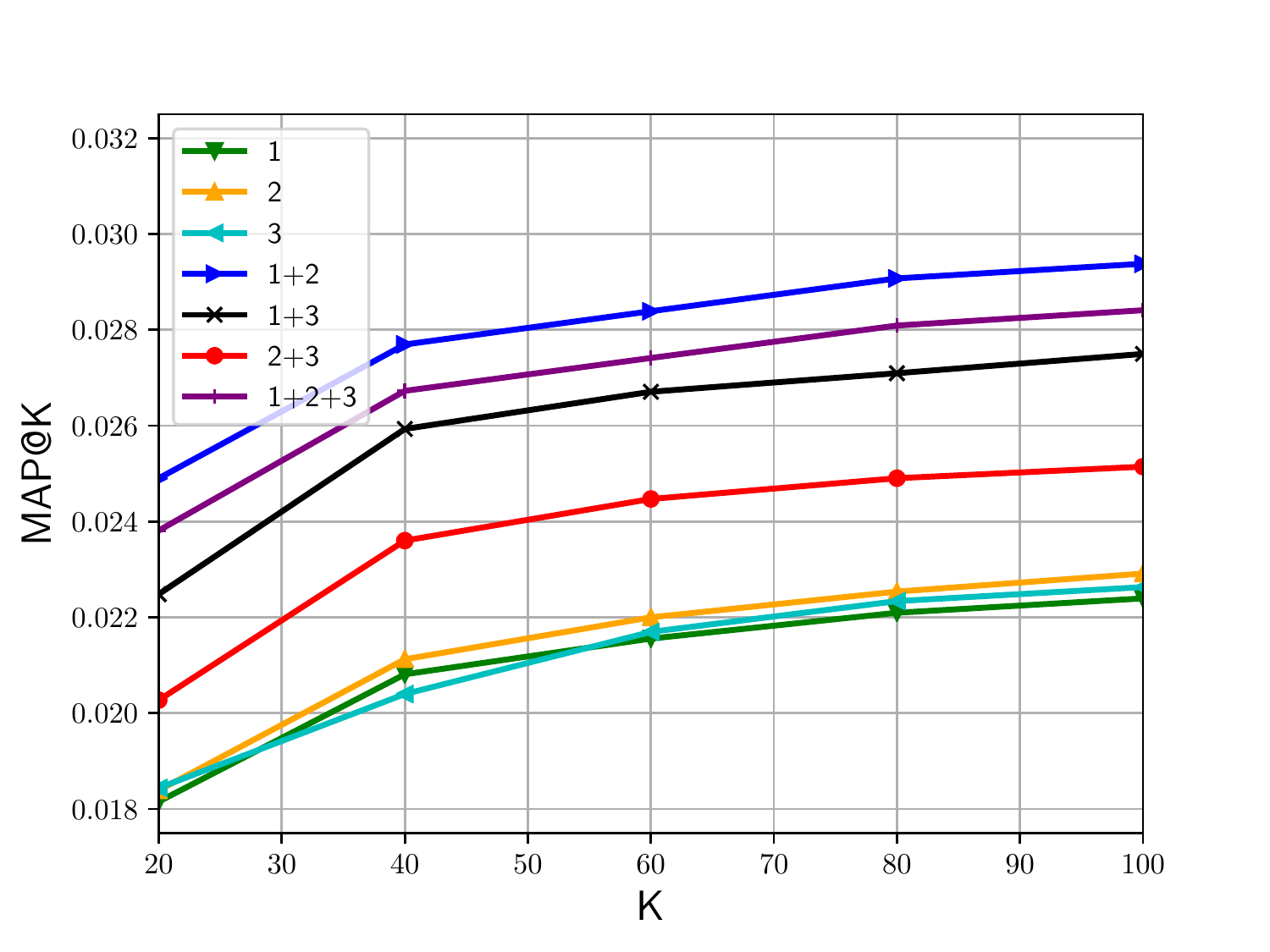}
    \vspace{-1em}
    \caption{Perfomrance of JSCN-$\beta$ w.r.t. MAP@K with different source domains and their combinations. The domain label can be referred in Table ~\ref{tab:multi-source domain}}
    \vspace{-1em}
    \label{fig:MAP_multi_source}
\end{figure}

We select three source domains: \textit{Home and Kitchen}, \textit{Health and Personal Care}, and \textit{Office Products}, which perform worst compared with the other source domains. The domain statistics are summarized in Table~\ref{tab:multi-source domain}. We conduct the experiment by choosing two out of three source domains to jointly learn the JSCN model. Hence we have $M=2$ and $M=3$ in this experiment. The comparison result is presented in Fig.~\ref{fig:MAP_multi_source}.

From the result, when $M=2$ we can find that multiple source domains can improve the performance compared with single source domain. Especially the combination of \textit{Home and Kitchen} and \textit{Health and Personal Care} source domains improve the performance by \textbf{37.2\%} on average compared using each one of the two domains. This experiment can prove the JSCN can jointly learn the information from multiple source domains. When $M=3$ we find the performance is a little bit worse than the combination of source domain 1 and source domain 2 (but still better than the other two combinations), which suggests that $M$ also requires tuning. The reason why $M=2$ is better than $M=3$ is that the source domain $3$ has smaller density value~\footnote{Density: 1 Home and Kitchen : $0.33\%$, 2 Health and Personal Care : $0.42\%$ and
3 Office Products : $0.14\%$ } compared with the other 2 domains, which can induce more disturbance to the model.

\subsection{Domain Adaptive Module}~\label{sec:domain adaptive}
In this part, we compare the performance of JSCN-$\alpha$ and JSCN-$\beta$ with different mapping functions. Recall that JSCN-$\alpha$ is the simple version of the joint spectral convolutional network which enforces the common user latent vector to be similar without the domain adaptive module. As for the domain adaptive module of JSCN-$\beta$, we have either the linear mapping or non-linear multi-layer perceptron~(MLP) mapping. We use four source domains, i.e. \textit{Books}, \textit{Movies and TV}~(MT), \textit{Clothing, Shoes and Jewelry}~(CSJ) and \textit{Apps for Android}~(AfA).

\begin{table}
\centering
\caption{The MAP@100 result of variants of JSCN}\label{tab:map variants of jscn}
\vspace{-1em}
\begin{tabular}{c|l|l|l|l}
\hline
 Source Domain & Books & MT & CSJ & AfA \\ 
 \hline
 JSCN-$\alpha$ & 0.02374 & 0.02291 & 0.02076 & 0.02103  \\
 JSCN-$\beta$-MLP & 0.02678 & \textbf{0.02375} & 0.02654 & 0.02537 \\
 JSCN-$\beta$ & \textbf{0.02769} & 0.02364 & \textbf{0.02877} & \textbf{0.03043}\\
  \hline
\end{tabular}
\end{table}

\begin{table}
\centering
\caption{The Recall@100 result of variants of JSCN}\label{tab:recall variants of jscn}
\vspace{-1em}
\begin{tabular}{c|l|l|l|l}
\hline
 Source Domain & Books & MT & CSJ & AfA \\ 
 \hline
 JSCN-$\alpha$ & 0.2011 & 0.2021 & 0.2050 & 0.2032  \\
 JSCN-$\beta$-MLP & 0.2107 & 0.2165 & 0.2112 & 0.2097 \\
 JSCN-$\beta$ & \textbf{0.2187} & \textbf{0.2179} & \textbf{0.2155} & \textbf{0.2217}\\
  \hline
\end{tabular}
\end{table}

From the result in Table~\ref{tab:map variants of jscn} and Table~\ref{tab:recall variants of jscn}, we can find JSCN-$\beta$ performs much better than JSCN-$\alpha$, which shows the effectiveness of the domain adaptive user mapping module. One interesting observation is the linear mapping beats the non-linear mapping. Since the non-linear mapping requires tuning a lot of hyper-parameters, such as choosing the activation function and the dimension of the hidden layer, we suggest using the linear mapping function for learning the invariant user vector. One possible explanation for this observation is that since the spectral vectors are already low dimensional vectors, MLP can easily find a mapping function such that the invariant user vectors in different domains to be the same, hence over-fitting the user vectors. As over-fitting will harm the structural regularization~\cite{zhou2011malsar} of the domain adaptive user mapping, the information cannot be transferred in a good way compared with linear mapping. 


\section{Conclusion}\label{Conclusion}
In this paper, we design a Joint Spectral Convolutional Network (JSCN) to solve the cross-domain recommendation problem. Firstly, JSCN operates multi-layer spectral convolutions on different graphs simultaneously. Secondly, JSCN maps the learned spectral latent vectors to a domain invariant user representation with adaptive user mapping module. Finally, JSCN minimizes both the in-domain loss in the spectral latent vector space and the cross-domain loss in the domain invariant user representation space to learn the parameters. From the experiment, we can answer three questions: 1)JSCN can use the source domain information to improve the recommendation performance; 2) the spectral convolutions in JSCN can capture the comprehensive connectivity information to improve the performance in cross-domain recommendation; 3) the adaptive user mapping of learning the domain-invariant representation can help to transfer knowledge across different domains. 

\section{Acknowledgement}
This work is supported in part by NSF under grants III-1526499, III-1763325, III-1909323, CNS-1930941, and CNS-1626432. This work is also partially supported by NSF through grant IIS-1763365 and by FSU.

\newpage

\bibliographystyle{IEEEtran}
\bibliography{references}


\end{document}